\documentclass{article}

\usepackage{tabularx}
\usepackage{float}
\usepackage[preprint]{neurips_2024}

\usepackage[utf8]{inputenc} 
\usepackage[T1]{fontenc}    
\usepackage{hyperref}       
\usepackage{url}            
\usepackage{booktabs}       
\usepackage{amsfonts}       
\usepackage{nicefrac}       
\usepackage{microtype}      
\usepackage{xcolor}         
\usepackage{amsmath}
\usepackage{graphicx} 
\usepackage{amsmath} 
\usepackage{amsthm} 
\usepackage{subfigure}
\theoremstyle{plain}

\theoremstyle{definition}

\theoremstyle{remark}

\title{Granger Causality Using Neural Networks}

\author{%
  Malik Shahid Sultan \\
  Department of Bio-engineering\\
  King Abdullah University of Science and Technology\\
  Thuwal, Saudi Arabia \\
  \And
  Samuel Horv\'{a}th \\
 Department of Machine Learning\\
 Muhammad Bin Zayed University of Artificial Intelligence \\
 Abu Dhabi, United Arab Emirates \\
  \And
  Hernando Ombao \\
 Department of Statistics \\
  King Abdullah University of Science and Technology\\
  Thuwal, Saudi Arabia \\
}

\begin{document}

\maketitle

\begin{abstract}
Dependence between nodes in a network is an important concept that pervades many areas including finance, politics, sociology, genomics and the brain sciences. One way to characterize dependence between components of a multivariate time series data is via Granger Causality (GC). Standard traditional approaches to GC estimation / inference commonly assume linear dynamics, however such simplification does not hold in many real-world applications where signals are inherently non-linear. In such cases, imposing linear models such as vector autoregressive (VAR) models can lead to mis-characterization of true Granger Causal interactions. To overcome this limitation, Tank et al (IEEE Transactions on Pattern Analysis and Machine Learning, 2022) proposed a solution that uses neural networks with sparse regularization penalties. The regularization encourages learnable weights to be sparse, which enables inference on GC. This paper overcomes the limitations of current methods by leveraging advances in machine learning and deep learning which have been demonstrated to learn hidden patterns in the data. We propose novel classes of models that can handle underlying non-linearity in a computationally efficient manner, simultaneously providing GC and lag order selection. Firstly, we present the Learned Kernel VAR (LeKVAR) model that learns kernel parameterized by a shared neural net followed by penalization on learnable weights to discover GC structure. Secondly, we show one can directly decouple lags and individual time series importance via decoupled penalties. This is important as we want to select the lag order during the process of GC estimation. This decoupling acts as a filtering and can be extended to any DL model including Multi-Layer Perceptrons (MLP), Recurrent Neural Networks (RNN), Long Short Term Memory Networks (LSTM), Transformers etc, for simultaneous GC estimation and lag selection. We use standard adaptive optimization algorithms that support mini-batching, to train our models instead of proximal gradient descent.The proposed techniques are evaluated on several benchmark simulation datasets.

\end{abstract}

\section{Introduction}
Precise forecasting is required in many applications including climate modeling \citet{mudelsee2019trend}, inventory management \citet{aviv2003time}, manufacturing  \citet{morariu2018time, 7005912} for data-driven decision making.
In these applications, past data is used to predict the future for data-driven planning and decision- making. However, many real data sets are marked by non-stationary and high volatility dynamics 
and thus it is a challenge to perform accurate forecasting. 
In neuroscience it is of paramount interest to study different brain regions and understand how the behavior evolves as a function of brain activity \citet{sporns2010networks,vicente2011transfer,stokes2017study,sheikhattar2018extracting}.
In such applications the prime objective of time series analysis is to uncover how to use the past in order to effectively and accurately predict the future. 
A range of various mathematical, statistical and machine learning models have been studied to forecast time series data. Popular choices for forecasting include vector autoregressive models \citet{fan2015wind}, generalized additive models \citet{serinaldi2011distributional}, random forests \citet{lin2017random}, XG-Boost \citet{wang2020forecasting}, support vector machines \citet{cao2003support} and deep neural networks \citet{torres2021deep}. There are many techniques and models in time and frequency domain that have been developed to capture the interactions within a time series and between different time series (e.g., auto-correlation, cross-correlation, cross-coherence, transfer information and directed entropy).

\subsection{Granger Causality}
Granger Causality (GC) \citet{granger1969investigating} is a statistical framework that can be used to test if the past of one time series is informative in the prediction of future of time series.
To understand the GC effect of a time series $X$ on a time series $Y$, under the two above mentioned assumptions, Granger proposed following hypothesis to test
\begin{align}
     \exists\; t \in \{1, 2, \hdots T\} \text{ s.t. }\mathbb{P}[Y(t+1)\in A\mid {\mathcal{X}}(t)]\neq \mathbb{P} [Y(t+1)\in A\mid {\mathcal{X}}_{-X}(t)],
\end{align}
where $\mathbb{P}$  refers to probability, $A$ is an arbitrary non-empty set, and $\mathcal {X}(t)$ and $\mathcal {X}_{-X}(t)$ respectively represents the data set of time $t$ with all of the variables, and that in the modified data set in which $X$ is excluded. If the above hypothesis is accepted, we say that $X$ Granger-causes $Y$. Bivariate analysis in presence of confounding (say time series $Z$) can be misleading therefore modeling GC as a network allows studying
high dimensional data \citet{lutkepohl2005new}. Traditionally GC has been extracted using coefficients of the vector autoregressive (VAR) models. However due to the linearity assumption these models may miss important structure in non-linear data. \citet{tank2018neural} relaxed the linearity constraint by using component wise NNs to model the non-linearity in data and infering GC from weights of input layer. 
In this work, we build on the previous works tackling GC on neural networks with the following key contributions. 

\begin{itemize}
    \item We propose a novel model to learn the non-linear dependence structure from the data, without specification of any kernel library. We model kernel function as a neural network which learns the shared kernel as a part of the training process. This proposed model is called the Learned Kernel Vector AutoRegressive (LeKVAR) model. 
    \item For Granger causality, one goal is to quantify the importance of each time series component and associated lag. The standard way to model GC using NN is to focus on the effect of each combination of lag and time-series component and then extract the importance of each time series component as a summation over lags. Here, we propose a simple and elegant approach to measure the significance of each lag and time series component by decoupling the lag and the time series. We include separate sparsity-inducing penalties for time series components and lags. We believe that such decoupling is less prone to over-fitting and provides a better estimate of GC. This also enables the user to easily incorporate prior knowledge as a mask or as a dynamic value of some regularization hyper-parameter. To the best of our knowledge, this proposed approach is the 
    first to enable direct lag extraction from recurrent neural models based on examination of learned model weights.
   \item We show that learning objective used to train models for estimation of GC happens to be degenerated and propose a solution. We propose a solution to overcome this issue by building the weight normalization into the learning. 
    \item In contrast to previous approaches \citet{tank2017granger,tank2018neural, khanna2019economy}, we train our models using standard DL optimizers, that enables mini-batching. Our solution is to directly incorporate penalty into the loss function that also allows us to use popular adaptive optimizers such as Adam~\citep{kingma2014adam} by avoiding the proximal step. We realize that such modification does not lead to exact zeros for GC, however thresholding scheme of \citet{marcinkevivcs2021interpretable} can be used based on time reversed GC.

\end{itemize}

\section{Related Works}
Popular methods used for GC estimation are classified as model-based or model-free approaches.
In model-based approaches, a parametric model is constructed (e.g., linear models such as the
VAR)
where the current value of a time series is modeled as a linear function of its own past values and the past of the other time series. Lag order selection is critical in these models, and the magnitude of the coefficients in these VAR models quantify the extent of GC: zero coefficients indicate a non-GC interaction between the time series. When the number of channels and the lag order are large, then the model becomes high dimensional; thus regularization constraints like Lasso \citet{tibshirani1996regression}, group lasso \citet{yuan2006model} are imposed on the parameters.
These sparsity-inducing penalties help to scale VAR models in high
dimensional settings. Linear assumption for estimation of GC can have serious implications when dealing with real-world data sets.
Therefore assuming a linear dependence for modeling the relationship between time series is an oversimplification and can lead to inconsistent and misleading results.
To relax the linearity assumption, \citet{sun2008assessing} used kernel functions to model the non-linearity. These non-linear basis functions can be used to model the non-linear dependence in the data set but rely heavily on the library of the kernel functions to be used. Therefore these kernel functions have to be inspired, motivated or driven by prior knowledge of the underlying process which are being modeled. Due to this constraint of fixed kernel functions and manual selection based on some domain- knowledge there is a chance that the specified kernel functions might not be suitable for the task or process under study. 
Another possibility is to use the Neural Networks (NN) to learn these non-linear transformations or kernel directly from the data without specifying any specific kernel in advance. We have dedicated a separate subsection entirely to learning the kernel in a data-adaptive manner using NNs. 
Model-based approaches can fail when model assumptions are not aligned with the underlying process. 
This can be observed as a result of the oversimplification of assumptions (e.g., fitting a linear 
model to a time series data whose underlying structure is non-linear). Transfer Entropy \citet{vicente2011transfer} and Directed Information \citet{amblard2011directed} are two 
techniques for model-free approaches that capture the dependence structure in the time series data 
with minimal assumptions and can effectively capture the non-linearity. However, one limitation of 
these approaches is the requirement to have large data to produce stable estimates. These methods also suffer from the curse of dimensionality when the number of the time series become large \citet{runge2012escaping}.
Autoregressive multi-Layer perceptrons (MLPs) \citet{raissi2018multistep,kicsi2004river,billings2013nonlinear}, Recurrent Neural Networks (RNNs) including Long Short Term Memory Networks (LSTMs) \citet{hochreiter1997long} have demonstrated high predictive power but are black-boxes. These methods have shown impressive predictive power, but are generally intrinsically not interpretable. 
The lack of interpretability of these models was studied in \citet{tank2018neural, tank2017granger}, which presented a framework of structure learning in MLPs and RNNs that leads to decoding and interpretation of the non-linear GC discovery. Their proposed framework has three main ingredients: (1) Leveraging the power of NNs to learn the complex relationships in the data; (2) Separate models for each time series to be predicted, they call this architecture as component wise cMLPs and component wise cLSTMs; (3) Imposing Sparsity inducing penalties on the input layer resulting in iterpretability of these models for GC discovery. Group sparsity penalties were used including group lasso and hierarchical group lasso \citet{yuan2006model, nicholson2014hierarchical, huang2011learning,kim2010tree} that serve two purposes: detection of GC and the identification of the lag structure. However, their method uses Proximal Gradient Descent for the optimization instead of adaptive DL optimizers. Economy Statistical Recurrent Units (eSRU) \citet{khanna2019economy} have also been proposed to infer GC from observed time series data, and are trained using approach similar to \citet{tank2018neural}. Self Explaining Neural Network for structure learning was proposed in \citet{marcinkevivcs2021interpretable} which uses NN to learn the weights for the coefficient matrix in VAR keeping the data fixed.


\section{LeKVAR}\label{chap:lekvar}
\begin{figure}
    \centering
    \includegraphics[width=0.5\textwidth]{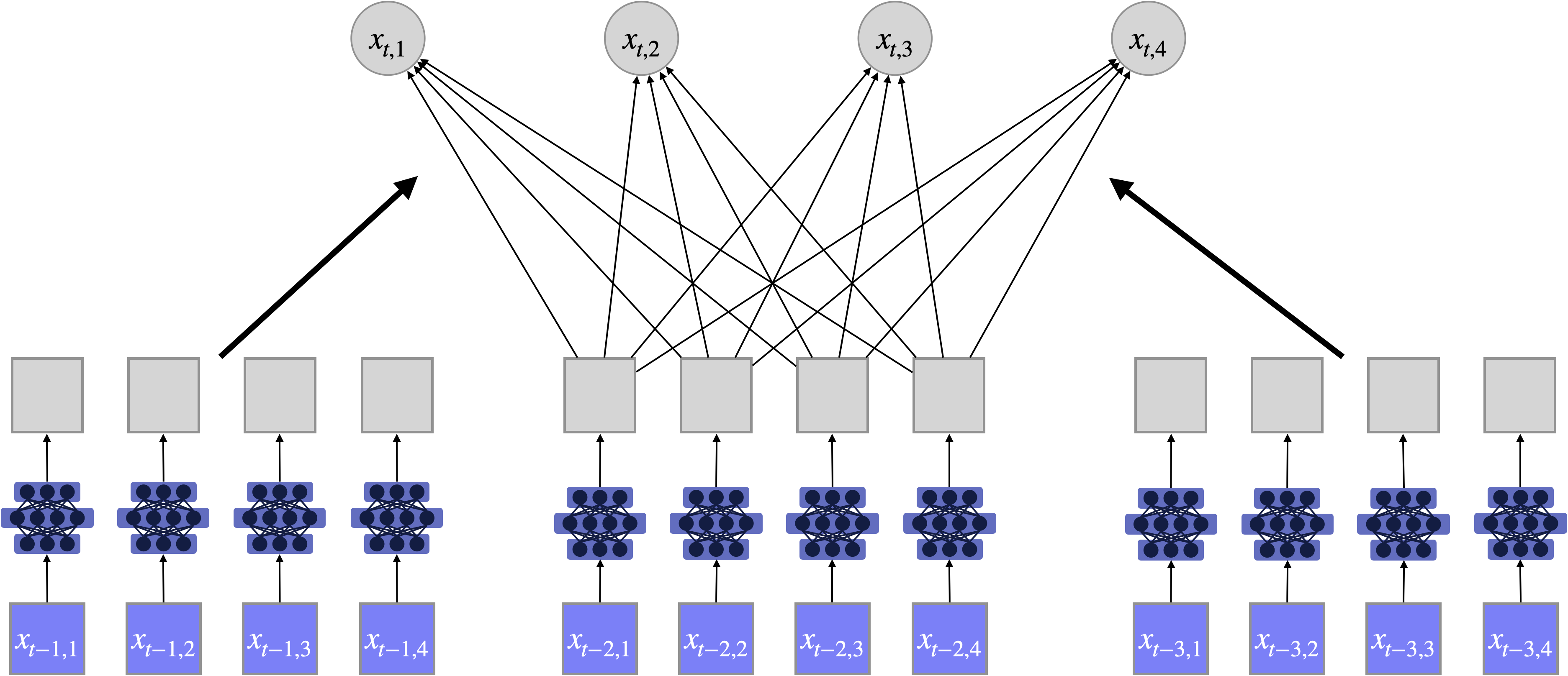}
    \caption{
    In LeKVAR the input is first transformed using a shared learned kernel displayed as a small purple neural network. Then each transformed input is linearly combined to form the final output. The round output neurons ($x_{t,i}$) represent the summation operation of incoming connections. We say that the series $j$ does not Granger-cause series $i$ if outgoing weights for series $j$, shown as dark arrows connecting gray squares with the output, are zeros. We say that the lag $h$ does not Granger-cause series $i$ if outgoing weights for series $[x_{t-h,1}, x_{t-h,2}, \hdots, x_{t-h,D}]$, shown as dark arrows connecting gray squares with the output, are zeros.}
    \label{fig:lekvar}
\end{figure}
This section proposes an extension of the GC VAR model that incorporates the learned kernel into the training process. Suppose $X$ $\in$ $R^{T \times D}$ is a $D$ dimensional time series for length $T$. We want to predict the future of ${\bf x}_t$ $\in$ $X$ using the information from the past $x_{t-h}$ $\in$ $B^{h}($X$)$ $h$ lags,where $B^{h}$ is the back shift operator. In our model, the time series at time $t$, ${\bf x}_t$, is assumed to be a linear combination of the past $H$ lags of the series which is transformed by a shared kernel 
\begin{equation}
{\bf x}_t = \sum_{h = 1}^H A^{(h)} \xi_{\theta}({\bf x}_{t-h}) + e_t,
\label{eq: LeKVar}
\end{equation}
where $\xi_{\theta}(\cdot)$ is the element-wise kernel function $\mathbb{R} \rightarrow \mathbb{R}$ parameterised by a neural network, with learnable parameters $\theta$. We call this as a single head. A kernel comprising of multiple heads can also be similarly learned. By enforcing kernel function to act independently on each element, we preserve the desired feature of VAR models that time series $j$ does not Granger-cause time series $i$ if and only if for all $h$, $A_{ij}^{(h)} = 0$. In case of a NN setting $A_{ij}^{(h)}$ would correspond to the weights for the final layer for node $j$ which is trained together with the $\xi$. Incorporating $\xi(\cdot)$ into the training process enables learning kernel instead of manual selection.
The schematic graph of the LeKVAR model is displayed in Figure~\ref{fig:lekvar}. 

Shared kernel functions have access to more data as the same kernel is shared across all components which allows for a more precise estimate and intuitively better performance when compared to a separate model for each time series component. This is computationally less expensive and scales better to higher dimensions when compared with component-wise models, as for a $D$ dimensional time series, instead of learning separate $D$ networks we now learn a single network, and infer GC from the output layer. In a NN setting, for a single time series, an analogue to component-wise models can be expressed as  $\ref{eq: LeKVar}$ as 
\begin{equation}
    {\bf x}^d_t =  f_{w^d}(\xi_{\theta}({\bf x}_{t-k})) + e_t,
\label{eq: LeKVar NN}
\end{equation}
here $f_{w^d}$ $\in$ $R^{H \times D}$ is the output layer transforming linearly the shared non-linear transformation. We want to minimize the following reconstruction loss along with regularization to extract GC time series and lags.

\begin{equation}
    \hat{f}_{w^d}, \hat{\xi}_{\theta} = \arg\min_{f_{w^d},\xi_{\theta}} ||x^{d} - \hat{x}^{d}||_{2}^{2} + \lambda ||w^{D}||_{F} + \lambda ||w^{H}||_{F}
\end{equation} 
where $w^{D}$ $\in$ $R^{D \times H \times D}$ and $w^{H}$ $\in$ $R^{H \times D \times D}$. The norm is calculated along the second dimension.
\section{Decoupling Lags and Time Series Components}\label{chap:decoupling}
\begin{figure}[t]
    \centering
    \includegraphics[width=0.65\textwidth]{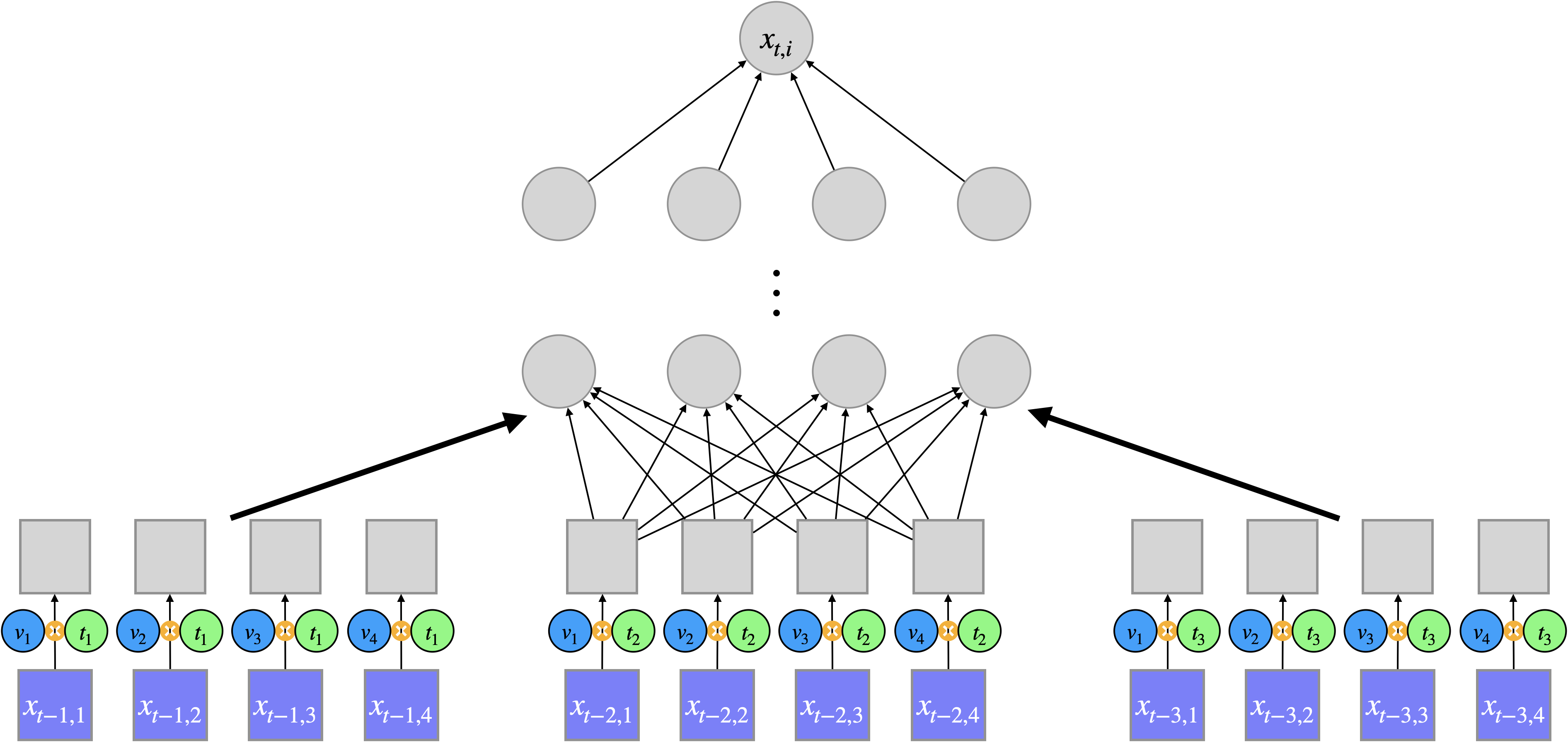}
    \caption{A schematic for the decoupling of the lags and times series components to decide Granger causality. Each input element is first multiplied by its corresponding lag and component penalty, i.e., $x_{t-h, d}$ is multiplied by the product $v_d t_h$, displayed as a small purple neural network. Then the output is obtained as the output of the component-wise neural network model with the scaled input. We say that the series $j$ does not Granger-cause series $i$ if the corresponding $v_j$ for the component-wise neural network to predict $x_{t, i}$, shown as a blue circle, is zero. We say that the lag $h$ does not Granger-cause series $i$ if the corresponding $t_h$ for the component-wise neural network to predict $x_{t, i}$, shown as a green circle, is zero.}
    \label{fig:regularization}
\end{figure}
Granger causality's main objective is to decide whether time series $j$ does not Granger-cause time series $i$. A typical approach to decide on Granger-cause of time series $j$ to time series $i$ is to look at the weights $W^{(i,j)}$ and compute its square norm. If this is less than some predefined threshold, we say time series $j$ does not Granger-cause time series $i$. For our new approach, we define vector $v^{i} \in \mathbb{R}^{2 \times D}$, which has two learnable components mean and variance.
$$v_{\mu} = (v^1_{\mu}, v^2_{\mu}, \hdots, v^D_{\mu})^\top$$

$$v_{\sigma^{2}} = (v^1_{\sigma^{2}}, v^2_{\sigma^{2}}, \hdots, v^D_{\sigma^{2}})^\top$$
to correspond to the mean importance of each time series and the variance around it.
Let $t^i \in \mathbb{R}^{2 \times H \times D}$ be an array that captures the importance of the lags corresponding to time series. This layer also has 2 components, a mean and a variance which are learnable.

$$t_{\mu} = (t^1_{\mu}, t^2_{\mu}, \hdots, t^D_{\mu})^\top$$
$$t_{\sigma^{2}} = (t^1_{\sigma^{2}}, t^2_{\sigma^{2}}, \hdots, t^D_{\sigma^{2}})^\top$$

In our experiments, we test two versions of this decoupling approach to estimate GC, one that involves only the mean component and another that involves both the mean and the variance. In case when we work with just the mean component the $v^{i} \in \mathbb{R}^{D}$ and  $t^{i} \in \mathbb{R}^{H \times D}$. 
To build this into the component-wise neural network model, we element-wise pre-multiply the data $X \in R^{BS \times H \times D}$ by $v^i$ followed by  $t^i$ and apply the sparsity-inducing penalties on $v^i$'s and $t^i$'s; see the schematic graph of the regularization model displayed in Figure~\ref{fig:regularization}. These multiplications can be implemented efficiently for standard machine learning libraries such as PyTorch \citep{paszke2017automatic}. This can be achieved by learning 2 element wise layers $V$ and $T$ in NNs that filter the data before passing it to the NN, we refer to the models built using this scheme as with filters $(WF)$ and $(WVF)$. Furthermore, we note that our approach can be naively combined with any sparsity inducing penalty.  
\begin{equation}
    x^d_{t} = g^{d}_{\theta}( t^d ( v^d (x_{t-H}))) + e^d_{t}
\end{equation}

\begin{equation}
    \arg\min_{g^{d}_{\theta},t^d,v^d} \frac{1}{n}||x^{d} - \hat{x}^{d}||_{2}^{2} + \lambda|v^d| + \lambda|t^d|
\end{equation}

where $g^{d}_{\theta}$ is any NN architecture like MLP, CNN, LSTM, RNN, GRU, Transformers etc. $t^d(.)$, and $v^d(.)$ needs special attention. In case of $t^d \in R^{H \times D}$ and $v^d \in R^{D}$, $t^d(.)$, and $v^d(.)$ are just element wise multiplication layers. 
To decide whether time series $j$ does not Granger-cause time series $i$, we look at the magnitude of $v^i_j$. Similarly the weights in array $t^d$ can be used to select the lags. $\lambda$ is the tune-able parameter and controls the sparsity, $|.|$ corresponds to L1 norm. In our experiments we perform 2 fold cross validation and grid search to select optimal $\lambda$ using time reversed GC procedure of \citet{marcinkevivcs2021interpretable}.

In case where $t^d \in R^{2 \times H \times D}$ and $v^d \in R^{2 \times D}$, $t^d(.)$, $v^d(.)$ are learning two components the mean and the variance. We employ reparameterization trick to first sample from the Gaussian distribution with these learnable parameters, followed by element wise multiplication operations. Let $p^{i}$ $\in$ $R^{D}$ and $q^{i}$ $\in$ $R^{H \times D}$.
\begin{equation}
    p^{i}_{j} = v^{i}_{\mu_{j}} + (z^{i}_{j} \times v^{i}_{\sigma_{j}})
\end{equation}

\begin{equation}
    q^{i}_{h,j} = t^{i}_{\mu_{h,j}} + (z^{i}_{h,j} \times t^{i}_{\sigma_{h,j}})
\end{equation}
where $z^{i}_{j}$ $\sim$ $N(0,1)$, therefore $p^{i}_{j}$ $\sim$ $N(v^{i}_{\mu_{j}},v^{i}_{\sigma^{2}_{j}})$ and $q^{i}_{h,j}$ $\sim$ $N(q^{i}_{\mu_{h,j}},q^{i}_{\sigma^{2}_{h,j}})$.

\begin{equation}
    x^d_{t} = g^{d}_{\theta}( q^d ( p^d (x_{t-H}))) + e^d_{t}
\end{equation}
\begin{equation}
    \arg\min_{g^{d}_{\theta},t^d,v^d} \frac{1}{n}||x^{d} - \hat{x}^{d}||_{2}^{2} + \lambda||v^d||_2^{2} + \lambda||t^d||_2^{2} + \gamma \sum_{i=1}^{D} KL(\psi || v^d_{i}) + \gamma \sum_{j=1}^{D} \sum_{h=1}^{H} KL(\psi || t^d_{h,j}) 
\end{equation}
where $\psi$ $\sim$ $N(0,\sigma^2_{\psi})$, $\sigma^2_{\psi}$ is a very small number close to zero.
To decide whether time series $j$ does not Granger-cause time series $i$, we observe at the magnitude of $v^i_{\mu_{j}}$ if it is close to zero. Similarly the weights in array $t^d$ can be used to select the lags. $\lambda$  and $\gamma$ are the tune-able parameter and controls the sparsity, $KL(||)$ corresponds to KL divergence. Adding $KL(||)$ is helpful as it helps in thresholding the weights for GC.

This procedure decouples the neural network model and Granger causality and provides better scaling regarding the number of parameters that need to estimated to infer GC connections. Finally, our approach allows us to capture the importance of each lag, and it is model agnostic. Hence, we can apply this to recurrent neural network (RNN) blocks that have not been possible before due to shared weights across different time steps. Furthermore, our proposed framework can estimate any dependence structure and build in prior knowledge by imposing structure in $v^d$'s or $t_d$'s without the necessity to tailor GC estimates to particular models. We show this by extending the approach to transformer models which has not been explored yet in literature.

\section{Degenerated Objective}\label{chap:degenerative}

We identify that the previous approaches to neural Granger causality including our proposed models (LeKVAR and decoupling) can't be naively applied as the learning objective happens to be degenerated. The issue is that we might multiply two parameters, where one of them is part the sparsity-inducing penalty and the other one is not, with each other without only scaling invariant transformation in between (e.g., no transformation or ReLU activation), which can lead to undesired behaviour. Let us demonstrate the problem via the following example.
We are interested in minimizing the following objective 
\begin{equation}
\label{eq:degenerated}
    \frac1n\sum_{i=1}^n \|v_1 w_1^\top x^i_1  + v_2 w_2^\top x^i_2 - y^i  \|_2^2 + \lambda (|v_1| + |v_2|)
\end{equation}
parameterized by scalars $v_1, v_2$ and vectors $w_1, w_2$, where $\lambda > 0$, $\{x^i_1, x^i_2\}_{i=1}^n$ are given vectors and $\{y^i\}_{i=1}^n$ are given scalars.  We note that for such objective penalty can be ignored as for any non-zero scalar $c < 1$ $v^{\prime}_1 = c v_1$, $v^{\prime}_2 = c v_2$ and $w^{\prime}_1 = \nicefrac{1}{c}\; w_1$, $w^{\prime}_2 =\nicefrac{1}{c}\; w_2$ lead to the same value of the first term while decreasing the second term. Therefore, we propose following reparameterization that allows us to incorporate penalty such that it has an actual effect on sparsity of the model. 
\begin{equation}
\label{eq:degenerated_fix}
    \frac1n\sum_{i=1}^n \left\|v_1 \frac{w_1^\top x^i_1}{\|w_1\|_2}  + v_2 \frac{w_2^\top x^i_2}{\|w_2\|_2} - y^i  \right\|_2^2 + \lambda (|v_1| + |v_2|).
\end{equation}
This reparametrization proceeds by normalizing weights connected to the penalty term, and it is an essential building block of our proposed methods.
One can think that such reparametrization might make optimizing the objective harder. Still, as investigated in \citep{salimans2016weight}, normalizing weights of the network can lead to better-conditioned problems and, hence, faster convergence. 
Note that in case ReLU is used as the activation function the weight normalization has to be applied for each layer. 

\section{Optimizing the Penalized Objectives}
\label{chap:optim}

A previously common approach to optimize the penalized objective was to run proximal gradient descent \citep{parikh2014proximal}. Proximal optimization was considered necessary in the context of GC because it leads to exact zeros. We propose embedding the sparsity-inducing penalty directly into the loss function following the standard training loop commonly considered in deep learning optimization. As such, it enables a wide variety of optimization tricks such as mini-batching, different forms of layer and weight normalization and usage of the popular adaptive optimizers, including Adam~\citep{kingma2014adam}, which are not naively compatible with proximal steps.

\begin{equation}
\min _{x \in \mathbb{R}^d} \frac{1}{n} \sum_{i=1}^n\left[f_i(x)+\lambda (|g_{v}(x)| +  |g_{t}(x)|) \right]
\end{equation}
Where $g_{v}$ and $g_{t}$ are the corresponding element wise learnable filtering layers and $\lambda$ controls the sparsity.
We realize that such modification does not lead to exact zeros for GC anymore. Still, in our experience, incorporating standard optimization tricks outweigh the issue of no exact zeros. We found it relatively easy to select the threshold by examining obtained GC coefficients in our experiments, and running stability based time reversed GC thresholding proposed in \citet{marcinkevivcs2021interpretable}.

\section{Experiments}\label{chap:experiments}
Our implementation will be made public with the camera-ready version of this article. We consider three tasks, namely the VAR model, Lorenz-96 Model and Netsim fMRI bold data set. All experiments were performed on $2\times$ Tesla V100 PCIE , 32768 MiB. We compare the models with , VAR, GVAR \cite{marcinkevivcs2021interpretable}, and eSRU \cite{khanna2019economy}. GC inference is perfomed using a 2 fold grid search procedure as used in \cite{marcinkevivcs2021interpretable}. The transformer based models cTRF and cVTRF, comprise of standard transformer encoder, a learnable positional encoding layer followed by a fully connected layer.

\subsection{VAR(3)}
We conclude a preliminary experimental study to assess how well can neural networks model perform on the relatively simple task of fitting 10-dimensional VAR(3), i.e., linear model, with $20\%$ of time series components being causal. Considering the performance of each method, we include a linear VAR model to provide an upper bound of how good the performance could be, F-test was used to obtain p-values using $\alpha = 0.05$ as the significance level using Benjamini-Hochberg \citet{benjamini1995controlling} procedure for controlling false discovery rate. The results for the simulation are given in table \ref{tab:VAR(3)}. We observe that the propose models perform on par with the baselines. The variational penalization helps to better estimate the GC when recurrent neural network or transformer based model is used. This can indicate that these models are more prone to over-fitting therefore adding coarse regularization may help to discover the GC structure in the data. LeKVAR model performs equally well with the best models in this scenario. Because here the true underlying kernel is linear therefore the LeKVAR does not over fit the data and is able to discover the true GC. This is important as LeKVAR does not require any kernel specification and is computationally efficient as well.

\begin{table}[h]
    \centering
\caption{Comparison of AUROC and AUPR for Granger causality selection among 10 dimensional VAR(3) model. Results are the mean across five random initialization, with one standard deviation}
\label{tab:VAR(3)}
    \begin{tabular}{|c|c|c|} \hline 
         Model&  AUROC& AUPRC\\ \hline 
         VAR&  0.97 $\pm$ 0.04& 0.87 $\pm$ 0.1\\ \hline 
         LeKVAR&  1.0 $\pm$ 0.0& 1.0 $\pm$ 0.0\\ \hline 
         GVAR&  1.0 $\pm$ 0.0& 1.0 $\pm$ 0.0\\ \hline 
         cMLPwF &  1.0 $\pm$ 0.0& 1.0 $\pm$ 0.0\\ \hline 
         cMLPwVF &  1.0 $\pm$ 0.0& 1.0 $\pm$ 0.0\\ \hline 
         cLSTMwF&  0.97 $\pm$ 0.05& 0.96 $\pm$ 0.07\\ \hline 
         cLSTMwVF&  0.99 $\pm$ 0.0009& 0.99 $\pm$ 0.008\\ \hline
         cTRF& 0.97 $\pm$ 0.03&0.92$\pm$ 0.04\\\hline
         cVTRF& 0.99 $\pm$ 0.02&0.97 $\pm$ 0.04\\\hline
    \end{tabular}

\end{table}

\subsection{Lorenz 96 Simulation}
A $p$-dimensional Lorenz model is characterized by continuous dynamics which can be stated as follows

\begin{equation}
\label{eq:lorenz}
\frac{d x_{t i}}{d t}=\left(x_{t(i+1)}-x_{t(i-2)}\right) x_{t(i-1)}-x_{t i}+F ,
\end{equation}
where $x_{t(-1)}=x_{t(p-1)}, x_{t 0}=x_{t p}, x_{t(p+1)}=x_{t 1}$ and $F$ is a
forcing constant that characterizes the degree of nonlinearity and chaotic behaviour in the series. In this study we conduct numerical simulations in the setting with parameters $p = 20$ Lorenz-96 model, sampling rate of $\Delta_{t}=0.05$ and $F=20$, which results in a multivariate, with a nonlinear behaviour, these time series exhibit a sparse Granger causal connections among its components.

\begin{table}[h]
    
    \centering
    
\caption{Comparison of AUROC and AUPRC for Granger causality selection among 20 dimensional Lorenz(20) model. Results are the mean across five random initialization, with one standard deviation}
\label{tab:lorenz}
    \begin{tabular}{|c|c|c|} \hline 
         Model&  AUROC& AUPRC\\ \hline 
         VAR&  0.75 $\pm$ 0.012& 0.54 $\pm$ 0.03\\ \hline 
         LeKVAR&  0.95 $\pm$ 0.02& 0.88 $\pm$ 0.03\\ \hline 
         GVAR&  0.99 $\pm$ 0.0006& 0.99$\pm$ 0.0003\\ \hline 
         cMLPwF&  0.95 $\pm$ 0.007& 0.87$\pm$ 0.03\\ \hline 
         cMLPwVF&  0.95$\pm$ 0.01& 0.85 $\pm$ 0.03\\ \hline 
 cLSTMwF& 0.99 $\pm$ 0.0003&0.99 $\pm$ 0.001\\ \hline 
 cLSTMwVF& 0.99 $\pm$ 0.003&0.95 $\pm$ 0.012\\ \hline 
 cTRF& 1.0 $\pm$ 0.0&1.0 $\pm$ 0.0\\ \hline
 cVCTRF& 0.98 $\pm$ 0.004&0.93 $\pm$ 0.02\\\hline

    \end{tabular}

\end{table}

We test our models' ability to recover the underlying causal structure. For Lorenz-96 model simulated data, average values of area under the ROC curve (AUROC) for recovery of the causal structure across 5 replicates are shown in table \ref{tab:lorenz}. The proposed component wise transformer model outperforms every other model in this simulation, however performance of all models except linear VAR is compareable. This simulation is characterized by high degree of non-linearity and we expect VAR to perform poorly here as compared to NN based methods. LeKVAR model has compareable performace with cMLPwF and cMLPwVF. Sequence based models including cLSTMwF,  cLSTMwVF, and cTRF and cVTRF have comparable performance. cVTRF and cLSTMwVF show a slight decay in AUROC and AUPRC when compared with the cTRF and cLSTMwF respectively, one reason for this can be small number of iterations that were performed to train these models. We perform 200 epochs therefore with less iterations less sampling when variational penalization is considered.


\subsection{Simulated fMRI Time Series}
The last dataset we consider to evaluate the performance of our proposed methods is the Blood Oxygen level Dependent (BOLD) time series \citet{smith2011network}. These realistic time series dataset was generated using functional magnetic resonance (fMRI) forward model. In our simulation we consider 5 data replicates from the simulation number 3, these are the same replicates which were also used in \cite{marcinkevivcs2021interpretable}. This multi-variate time series data has 15 dimensions and $T = 200$ time points. Therefore this is a data scarce scenario. The ground tructh GC is very sparse. We provide the comparison of the methods in table \ref{tab:fMRI-Netsim}. We observe that LSTMwVF has the best performance. The performance of models with variational penalization is comparable with counterparts with only L1 regularization. LeKVAR model performs better than GVAR and is also computationally less expensive. This shows that LeKVAR can also be used in non-linear and data scarce scenarios, with the benefit of learning the kernel from the data rather then specifying any kernel from prior knowledge.
\begin{table}[h]
    \centering
\caption{Comparison of AUROC and AUPRC for Granger causality selection among 15 dimensional
simulated Netsim fMRI Bold time series dataset. Results are the mean across five replicates, with one standard deviation}
\label{tab:fMRI-Netsim}
    \begin{tabular}{|c|c|c|} \hline 
         Model&  AUROC& AUPRC\\ \hline 
 VAR& 0.50$\pm$ 0.009&0.08 $\pm$ 0.003\\ \hline 
 LeKVAR& 0.68 $\pm$ 0.09&0.26 $\pm$ 0.19\\\hline 
         GVAR&  0.65 $\pm$ 0.024& 0.22 $\pm$ 0.05\\ \hline 
 eSRU& 0.65 $\pm$ 0.06 &0.19 $\pm$ 0.02\\ \hline 
 cMLPwF& 0.72 $\pm$ 0.07&0.33 $\pm$ 0.05\\ \hline 
         CMLPwVF&  0.73 $\pm$ 0.08& 0.31 $\pm$ 0.08\\ \hline
 LSTMwF& 0.77 $\pm$ 0.06&0.31 $\pm$ 0.08\\\hline
 LSTMwVF& 0.78 $\pm$ 0.05&0.3203 $\pm$ 0.08\\\hline
 cTRF& 0.7156 $\pm$ 0.04&0.22 $\pm$ 0.04\\\hline
 cVTRF& 0.69 $\pm$ 0.01&0.26 $\pm$ 0.06\\\hline
    \end{tabular}

\end{table}%

\subsection{Comparison of Computation Times}
We compare the computation time needed for the forward pass for all of the models under study with 2 hidden layers with 100 neurons each. We observe that recurrent models are the most expensive computationally. LeKVAR is at least 30x faster than compared to LSTMs and 3x faster then GVAR. This helps us show that the proposed LeKVAR has computationally superior performance as compared to other models understudy with a comparable performance in terms of estimating GC.
\begin{table}[t]
    \centering
\caption{Comparison of computation time for a forward pass for a 100 dimensional data set with batch size 256 and 10 lags . All networks have 100 hidden units with 2 layers. The times are average over 100 random initialization}
\label{tab:my_label}
    \begin{tabular}{|c|c|c|} \hline 
         Model& Average Computation Time (sec) & Ratio w.r.t to LeKVAR\\ \hline 
         GVAR& 0.3956 &3.1125\\ \hline 
         CMLPwF& 0.3081 &2.4240\\ \hline 
         CMLPwVF& 0.3236 &2.5460\\ \hline 
         cTRFwF& 3.4841 &27.4122\\ \hline 
         cTRFwVF& 3.5415 &27.8638\\ \hline
 LSTMwF&4.511 &35.4925\\\hline
 LSTMwVF&4.3744 &34.4169\\\hline
 LeKVAR&0.1271 &1\\\hline
    \end{tabular}

\end{table}

\section{Conclusion and Future Work}

This paper looks at the different approaches to estimating Granger causality (GC) using neural network models. We start by revising the prior work and identifying several advantages and disadvantages of the previously proposed models. 
We propose LeKVAR as an extension to the kernel methods where we parameterize the kernel using a neural network model. 
We include its learning as a part of the training procedure. 
 Furthermore, we develop a new framework that allows us to decouple model selection and Granger causality that provides a lot of promising features, e.g., lag estimates for RNNs and build-in prior knowledge, leading to improved performance in inferring the Granger-causal connections. Next, we identify and fix the issue that the previous approaches to neural Granger causality, including our proposed models, can't be naively applied as the learning objective is degenerated, i.e., it does not promote sparsity regardless of the sparsity parameter. We also propose a new problem reformulation that enables us to use popular tricks to optimize neural networks. 
 We evaluate the proposed methods on simulated toy and real-world datasets, and we conclude that NN-based models are compatible and can be used to infer Granger causality. We show that decoupling helps us identify GC interactions and lags simultaneously and can easily be extended to any NN architecture including transformers. However, efficient tuning/optimization techniques remain a formidable challenge that we want to tackle in future work.

\bibliographystyle{plainnat}
\bibliography{main.bib}
\newpage
\section{Appendix}

\subsection{Experiment Details}
\subsubsection{VAR(3) model}
\begin{figure}[H]
    \begin{subfigure}[b]
        \centering
        \includegraphics[width=0.4 \textwidth]{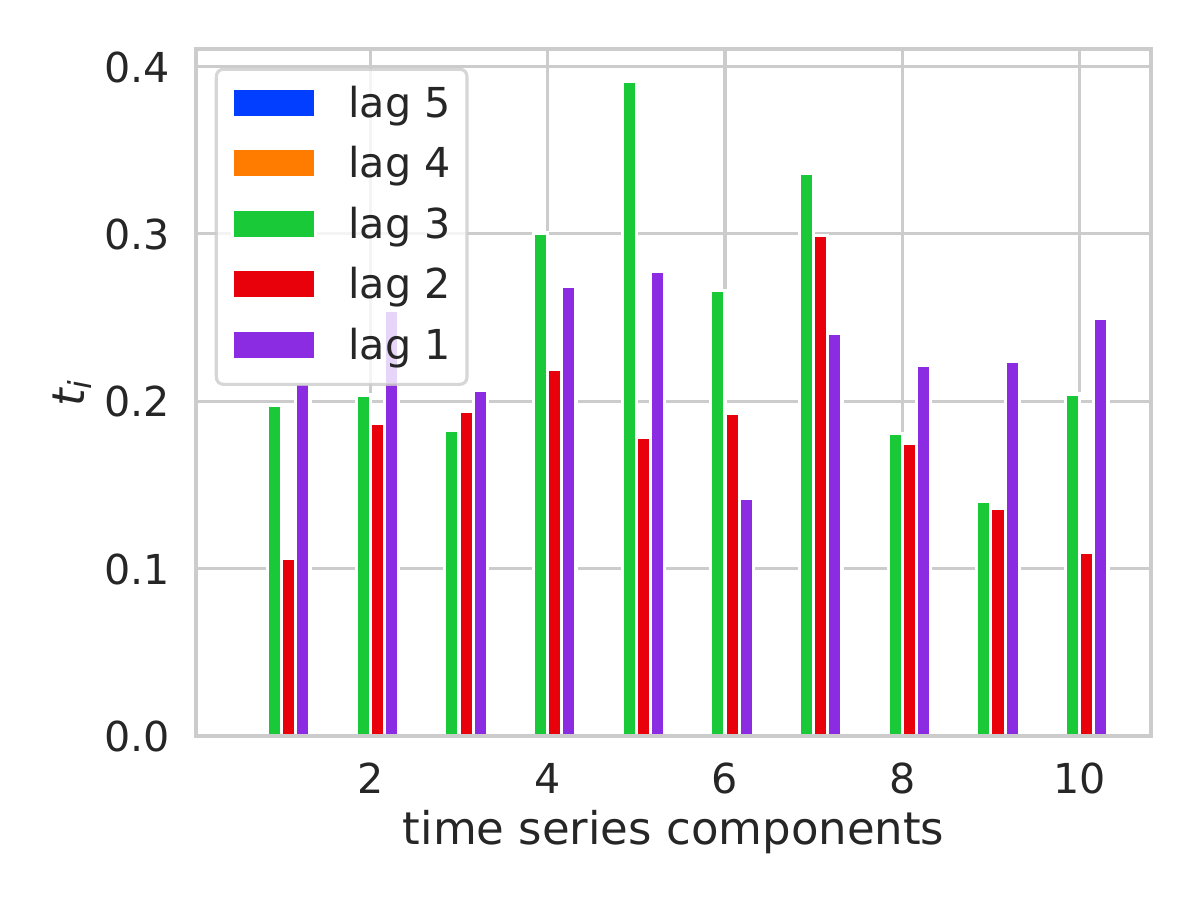} 
        \includegraphics[width=0.4 \textwidth]{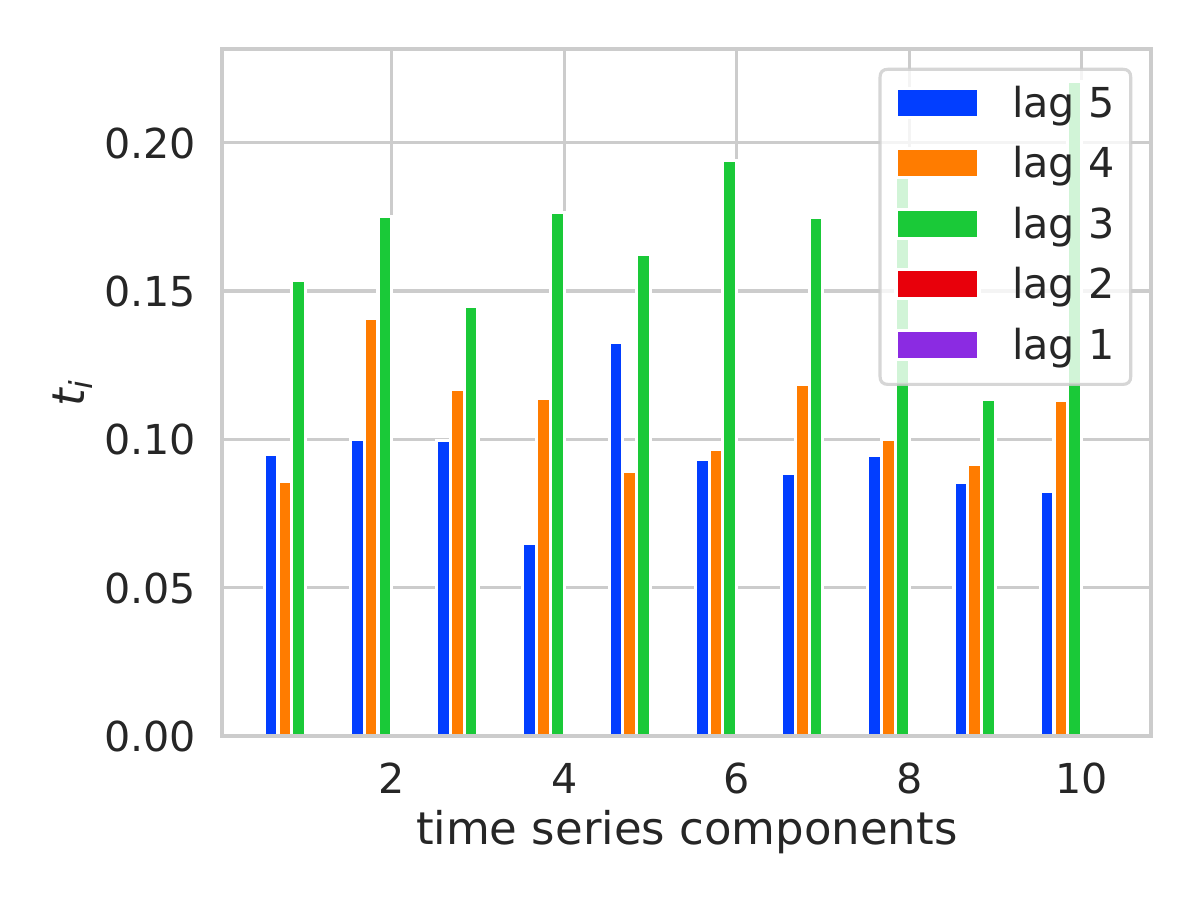}

    \end{subfigure}
    \caption{Comparison of learned Granger causality coefficients $t_i$'s of 10 dimensional VAR dataset using the cLSTMwF model. (Left) Ground truth lags are $1, 2$ and $3$, (right) ground truth lags are $3, 4$ and $5$. These lags information can only be extracted from $cLSTMwf$ models not from $cLSTM$}
    \label{fig:lags}
\end{figure}

To showcase that we can extract lags in recurrent neural network with our new framework introduced in Section~\ref{chap:degenerative}, we displayed obtained $t_i$'s coefficient for the cLSTMwF model. To further showcase flexibility, we also generate data where the lags 3, 4 and 5 are causal. Figure~\ref{fig:lags} displays the obtained $t_i$'s coefficient and shows that it is indeed the case that for the sufficiently large number of data, our approach is able to exactly recover the underlying lag structure.This shall be noted here that this structure cannot be naively discovered with RNNs without using our method of hierarchical filtering.

\begin{table}[H]
    \centering
    \small 
    \begin{tabularx}{\textwidth}{|c|c|c|c|X|c|} \hline 
         Model & Hidden Units & $\#$ of Hidden Layers & Batch Size & Hyper-Parameters & Lags H \\ \hline 
         GVAR & 50 & 2 & 64 & $\lambda \in 0 - 3, \gamma \in 0 - 0.025$ & 10 \\ \hline 
         LeKVAR & 50 & 2 & 64 & $\lambda \in 1 - 1e-7$ & 10 \\ \hline 
         cMLPwF & 50 & 2 & 64 & $\lambda \in 1 - 1e-7$ & 10 \\ \hline 
         cMLPwVF & 50 & 2 & 64 & $\lambda \in 1 - 1e-7$ & 10 \\ \hline 
         LSTMwF & 50 & 1 & 64 & $\lambda \in 1 - 1e-7$ & 10 \\ \hline 
         LSTMwVF & 50 & 1 & 64 & $\lambda \in 1 - 1e-7, \gamma \in 1 - 1e-7$ & 10 \\ \hline 
         cTRF & 50 & 1 & 64 & $\lambda \in 1 - 1e-7, \gamma \in 1 - 1e-7$ & 10 \\ \hline 
         VCTRF & 50 & 1 & 64 & $\lambda \in 1 - 1e-7, \gamma \in 1 - 1e-7$ & 10 \\ \hline
    \end{tabularx}
    \caption{Hyper-parameter Settings for VAR(3)}
    \label{tab:Hyper-Params VAR (3)}
\end{table}

\begin{table}[H]
\caption{Comparison of AUROC and AUPR for Granger causality selection among 10 dimensional VAR(3) model.
Results are the mean across three random initialization, with one standard deviation.}
\label{tab:var_3_Complete}
\begin{tabular}{@{}crrrccc@{}}
\toprule
\multicolumn{1}{l}{}                      & \multicolumn{3}{c}{\textbf{AUROC}}                                                                                & \multicolumn{3}{c}{\textbf{AUPR}}                                         \\ \midrule
\multicolumn{1}{|c|}{\textit{\textbf{T}}} & \multicolumn{1}{c}{\textit{\textbf{100}}} & \multicolumn{1}{c}{\textbf{200}} & \multicolumn{1}{c|}{\textbf{1000}} & \textit{\textbf{100}} & \textbf{200} & \multicolumn{1}{c|}{\textbf{1000}} \\ \midrule
\multicolumn{1}{|c|}{\textbf{VAR}}        & 77.6 ± 1.2                                & 98.9 ± 0.1                       & \multicolumn{1}{c|}{100.0 ± 0.0}   & 49.4 ± 3.7            & 95.0 ± 0.0   & \multicolumn{1}{c|}{100.0 ± 0.0}   \\ \midrule
\multicolumn{1}{|c|}{\textbf{LeKVAR}}     & 65.1 ± 9.0                                & 99.5 ± 0.4                       & \multicolumn{1}{r|}{100.0 ± 0.0}   & 56.6 ± 6.8            & 98.2 ± 1.4   & \multicolumn{1}{c|}{100.0 ± 0.0}   \\ \midrule
\multicolumn{1}{|c|}{\textbf{cMLP}}       & 73.1 ± 5.4                                & 97.5 ± 1.8                       & \multicolumn{1}{r|}{100.0 ± 0.0}   & 53.7 ± 8.6            & 90.2 ± 7.0   & \multicolumn{1}{c|}{100.0 ± 0.0}   \\ \midrule
\multicolumn{1}{|c|}{\textbf{cMLPwF}}     & 69.5 ± 2.8                                & 97.5 ± 1.3                       & \multicolumn{1}{r|}{100.0 ± 0.0}   & 43.3 ± 8.8            & 91.4 ± 4.3   & \multicolumn{1}{c|}{100.0 ± 0.0}   \\ \midrule
\multicolumn{1}{|c|}{\textbf{cLSTM}}      & 74.9 ± 2.9                                & 97.9 ± 0.8                       & \multicolumn{1}{r|}{100.0 ± 0.0}   & 53.8 ± 5.2            & 94.2 ± 1.8   & \multicolumn{1}{c|}{100.0 ± 0.0}   \\ \midrule
\multicolumn{1}{|c|}{\textbf{cLSTMwF}}    & 73.6 ± 2.7                                & 99.6 ± 0.4                       & \multicolumn{1}{c|}{100.0 ± 0.0}   & 41.1 ± 6.1            & 98.9 ± 0.8   & \multicolumn{1}{c|}{100.0 ± 0.0}   \\ \midrule
\multicolumn{1}{|c|}{\textbf{cMLP\_s}}       & 79.2 ± 1.9                                & 98.6 
± 0.5                       & \multicolumn{1}{r|}{100.0 ± 0.0}   & 60.4 ± 4.6            & 93.7 ± 2.0   & \multicolumn{1}{c|}{100.0 ± 0.0}   \\ \midrule
\multicolumn{1}{|c|}{\textbf{cMLPwF\_s}}     & 76.3 ± 3.5                                & 97.5 ± 1.0                      & \multicolumn{1}{r|}{100.0 ± 0.0}   & 51.3 ± 7.0            & 88.6 ± 5.0   & \multicolumn{1}{c|}{100.0 ± 0.0}   \\ \midrule
\multicolumn{1}{|c|}{\textbf{cLSTM\_s}}      & 78.2 ± 2.2                                & 98.6 ± 0.5                       & \multicolumn{1}{r|}{100.0 ± 0.0}   & 62.0 ± 5.8            & 94.8 ± 2.6   & \multicolumn{1}{c|}{100.0 ± 0.0}   \\ \midrule
\multicolumn{1}{|c|}{\textbf{cLSTMwF\_s}}    & 74.1 ± 0.6                                & 98.1 ± 1.0                       & \multicolumn{1}{c|}{100.0 ± 0.0}   & 60.1 ± 2.2            & 94.6 ± 2.2   & \multicolumn{1}{c|}{100.0 ± 0.0}   \\ \bottomrule
\end{tabular}
\end{table}
We also provide the results for the simulations where we compare our models in VAR(3) simulation with different length of time series given in table \ref{tab:var_3_Complete}.

\subsubsection{Lorenz-96 Model}
We provide the details of the model settings in the table \ref{tab:Hyper-Params Lorenz (20)}. Apart from this we also provide the ablation study details where we study the performance of the models as a function of length of time series across 3 data sets generated with a forcing constant set to 20.

\begin{table}[H]
    \centering
    \small 
    \begin{tabularx}{\textwidth}{|c|c|c|c|X|c|} \hline 
         Model & Hidden Units & $\#$ of Hidden Layers & Batch Size & Hyper-Parameters & Lags H \\ \hline 
         GVAR & 50 & 2 & 64 & $\lambda \in 0 - 3, \gamma \in 0 - 0.025$ & 5 \\ \hline 
         LeKVAR & 50 & 2 & 64 & $\lambda \in 1 - 1e-7$ & 5 \\ \hline 
         cMLPwF & 50 & 2 & 64 & $\lambda \in 1 - 1e-7$ & 5 \\ \hline 
         cMLPwVF & 50 & 2 & 64 & $\lambda \in 1 - 1e-7$ & 5 \\ \hline 
         LSTMwF & 50 & 1 & 64 & $\lambda \in 1 - 1e-7$ & 5 \\ \hline 
         LSTMwVF & 50 & 1 & 64 & $\lambda \in 1 - 1e-7, \gamma \in 1 - 1e-7$ & 5 \\ \hline 
         cTRF & 50 & 1 & 64 & $\lambda \in 1 - 1e-7, \gamma \in 1 - 1e-7$ & 5 \\ \hline 
         VCTRF & 50 & 1 & 64 & $\lambda \in 1 - 1e-7, \gamma \in 1 - 1e-7$ & 5 \\ \hline
    \end{tabularx}
    \caption{Hyper-parameter Settings for Lorenz (20) F = 20}
    \label{tab:Hyper-Params Lorenz (20)}
\end{table}

\begin{table}[H]
\caption{Comparison of AUROC and AUPR for GC selection among 20 dimensional Lorenz-96 model.Results are the mean across three random initializations, with one standard deviation, parametrised as a function of the length of the time series $T$.} 
\label{tab:lorenz_COMPLETE}
\begin{tabular}{@{}crrrccc@{}}
\toprule
\multicolumn{1}{l}{}                      & \multicolumn{3}{c}{\textbf{AUROC}}                                                                                & \multicolumn{3}{c}{\textbf{AUPR}}                                         \\ \midrule
\multicolumn{1}{|c|}{\textit{\textbf{T}}} & \multicolumn{1}{c}{\textit{\textbf{250}}} & \multicolumn{1}{c}{\textbf{750}} & \multicolumn{1}{c|}{\textbf{1500}} & \textit{\textbf{100}} & \textbf{200} & \multicolumn{1}{c|}{\textbf{1000}} \\ \midrule
\multicolumn{1}{|c|}{\textbf{VAR}}        & 83.9 ± 0.4 & 96.7 ± 0.1 & \multicolumn{1}{c|}{99.1 ± 0.1} & 71.0 ± 0.3 & 93.2 ± 0.3 & \multicolumn{1}{c|}{97.0 ± 0.2}   \\ \midrule
\multicolumn{1}{|c|}{\textbf{LeKVAR}}     & 84.5 ± 0.6 & 94.8 ± 3.1 & \multicolumn{1}{r|}{100.0 ± 0.0} & 71.1 ± 2.8 & 91.5 ± 3.9 & \multicolumn{1}{c|}{99.9 ± 0.0}   \\ \midrule
\multicolumn{1}{|c|}{\textbf{cMLP}}       & 88.5 ± 0.4 & 97.6 ± 0.5 & \multicolumn{1}{r|}{99.9 ± 0.0} & 77.2 ± 0.5 & 95.2 ± 0.5 & \multicolumn{1}{c|}{99.8 ± 0.1}   \\ \midrule
\multicolumn{1}{|c|}{\textbf{cMLPwF}}     & 89.8 ± 1.3 & 94.7 ± 0.2 & \multicolumn{1}{r|}{90.4 ± 2.5} & 78.0 ± 1.5 & 92.6 ± 0.9 & \multicolumn{1}{c|}{84.9 ± 3.8}   \\ \midrule
\multicolumn{1}{|c|}{\textbf{cLSTM}}      & 50.8 ± 3.3 & 73.0 ± 2.3 & \multicolumn{1}{r|}{98.2 ± 0.6} & 22.8 ± 2.3 & 43.4 ± 2.1 & \multicolumn{1}{c|}{92.3 ± 2.4}   \\ \midrule
\multicolumn{1}{|c|}{\textbf{cLSTMwF}}    & 54.4 ± 3.9 & 91.6 ± 2.3 & \multicolumn{1}{c|}{97.9 ± 0.1} & 24.3 ± 2.7 & 76.7 ± 7.6 & \multicolumn{1}{c|}{90.6 ± 0.8}   \\ \midrule
\multicolumn{1}{|c|}{\textbf{cMLP\_s}}    & 92.0 ± 0.3 & 96.1 ± 0.8 & \multicolumn{1}{r|}{99.8 ± 0.1} & 79.6 ± 0.4 & 88.4 ± 4.1 & \multicolumn{1}{c|}{99.3 ± 0.4}   \\ \midrule
\multicolumn{1}{|c|}{\textbf{cMLPwF\_s}}  & 88.4 ± 1.4 & 99.0 ± 0.3 & \multicolumn{1}{r|}{99.5 ± 0.0} & 75.0 ± 3.7 & 97.0 ± 0.6 & \multicolumn{1}{c|}{98.0 ± 0.1}   \\ \midrule
\multicolumn{1}{|c|}{\textbf{cLSTM\_s}}   & 56.6 ± 3.5 & 96.6 ± 0.6 & \multicolumn{1}{r|}{99.1 ± 0.1} & 25.6 ± 4.2 & 91.0 ± 1.2 & \multicolumn{1}{c|}{96.4 ± 0.6}   \\ \midrule
\multicolumn{1}{|c|}{\textbf{cLSTMwF\_s}} & 58.3 ± 0.5 & 87.8 ± 1.2 & \multicolumn{1}{c|}{98.4 ± 0.2} & 25.8 ± 0.5 & 56.1 ± 4.2 & \multicolumn{1}{c|}{93.1 ± 1.0}   \\ \bottomrule
\end{tabular}

\end{table}

\subsubsection{fMRI BOLD Netsim Simulation}

\begin{table}[H]
    \centering
    \small 
    \begin{tabularx}{\textwidth}{|c|c|c|c|X|c|} \hline 
         Model & Hidden Units & $\#$ of Hidden Layers & Batch Size & Hyper-Parameters & Lags H \\ \hline 
         GVAR & 50 & 1 & 64 & $\lambda \in 0 - 3, \gamma \in 0 - 0.025$ & 1 \\ \hline \hline 
         LeKVAR & 50 & 1 & 64 & $\lambda \in 1 - 1e-7$ & 1 \\ \hline 
         cMLPwF & 50 & 1 & 64 & $\lambda \in 1 - 1e-7$ & 1 \\ \hline 
         cMLPwVF & 50 & 1 & 64 & $\lambda \in 1 - 1e-7$ & 1 \\ \hline 
         LSTMwF & 50 &  & 64 & $\lambda \in 1 - 1e-7$ & 1 \\ \hline 
         LSTMwVF & 50 & 1 & 64 & $\lambda \in 1 - 1e-7, \gamma \in 1 - 1e-7$ & 1 \\ \hline 
         cTRF & 50 & 1 & 64 & $\lambda \in 1 - 1e-7, \gamma \in 1 - 1e-7$ & 1 \\ \hline 
         VCTRF & 50 & 1 & 64 & $\lambda \in 1 - 1e-7, \gamma \in 1 - 1e-7$ & 1 \\ \hline
    \end{tabularx}
    \caption{Hyper-parameter Settings for fMRI Netsim}
    \label{tab:Hyper-Params Netsim}
\end{table}



\end{document}